\documentclass[letterpaper]{article} 
\usepackage{aaai25}  
\usepackage{times}  
\usepackage{helvet}  
\usepackage{courier}  
\usepackage[hyphens]{url}  
\usepackage{graphicx} 
\usepackage{natbib}  
\usepackage{caption} 
\usepackage{algorithm}
\usepackage{algorithmic}

\usepackage{newfloat}
\usepackage{listings}
\DeclareCaptionStyle{ruled}{labelfont=normalfont,labelsep=colon,strut=off} 
\floatstyle{ruled}
\newfloat{listing}{tb}{lst}{}
\floatname{listing}{Listing}
\usepackage{amssymb}
\usepackage{amsmath}
\usepackage{booktabs}
\usepackage{multirow}
\usepackage{subfig}

\title{FFCG: Effective and Fast Family Column Generation \\for Solving Large-Scale Linear Program}
\author{
    Yi-Xiang Hu\textsuperscript{\rm 1}, Feng Wu\textsuperscript{\rm 1*}, Shaoang Li\textsuperscript{\rm 1}, Yifang Zhao\textsuperscript{\rm 2}, Xiang-Yang Li\textsuperscript{\rm 1}\thanks{Corresponding authors.}
}
\affiliations{
  
\textsuperscript{\rm 1}School of Computer Science and Technology, University of Science and Technology of China\\ \textsuperscript{\rm 2}School of Cyber Science and Technology, University of Science and Technology of China\\
    \{yixianghu,lishaoa,zhaoyifang\}@mail.ustc.edu.cn,\{wufeng02,xiangyangli\}@ustc.edu.cn
}

\usepackage{bibentry}

\begin{document}

\maketitle

\begin{abstract}
Column Generation (CG) is an effective and iterative algorithm to solve large-scale linear programs (LP). During each CG iteration, new columns are added to improve the solution of the LP. Typically, CG greedily selects one column with the most negative reduced cost, which can be improved by adding more columns at once. However, selecting all columns with negative reduced costs would lead to the addition of redundant columns that do not improve the objective value. Therefore, selecting the appropriate columns to add is still an open problem and previous machine-learning-based approaches for CG only add a constant quantity of columns per iteration due to the state-space explosion problem. To address this, we propose Fast Family Column Generation (FFCG) --- a novel reinforcement-learning-based CG that selects a variable number of columns as needed in an iteration. Specifically, we formulate the column selection problem in CG as an MDP and design a reward metric that balances both the convergence speed and the number of redundant columns. In our experiments, FFCG converges faster on the common benchmarks and reduces the number of CG iterations by 77.1\% for Cutting Stock Problem (CSP) and 84.8\% for Vehicle Routing Problem with Time Windows (VRPTW), and a 71.4\% reduction in computing time for CSP and 84.0\% for VRPTW on average compared to several state-of-the-art baselines.
\end{abstract}

%

\section{Introduction}

In many real-world applications, the successful solution of large-scale Mixed-Integer Linear Programming (MILP) problems requires solving Linear Programming (LP) relaxations that have a huge number of variables. For example, in the Cutting Stock Problem (CSP) with a length of $n$, each integer variable represents the number of times a cutting pattern is used, and there are exponentially many ($2^n$) cutting patterns. However, LP involving a large number of variables (i.e., columns) often cannot be handled at once by the solver. 
To address this, \citet{Gilmore} proposed an iterative algorithm, called Column Generation (CG), which is widely used to solve large-scale LPs. 
Specifically, CG starts by solving a Restricted Master Problem (RMP) with a subset of columns and gradually includes new columns, which can improve the solution of the current RMP, by solving a pricing subproblem (SP). When no more columns with negative reduced cost are found, CG will provably converge to an optimal solution to the LP. In practice, CG is often used by LP relaxation solvers in the branch-and-price algorithm \cite{RePEc:spr:sprbok:978-0-387-25486-9} for large MILP. 

Here, we focus on how to select columns in each iteration efficiently and speed up the convergence of CG. Typically, CG greedily selects
a single column with the most negative reduced cost, which can be sped up by adding multiple columns at once \cite{ascg}.
To distinguish from single-column selection, we refer to CG with multi-column selection as Family Column Generation (FCG). 
Unfortunately, adding multiple columns raises another issue: if CG greedily selects all columns with negative reduced costs, many redundant columns that do not improve the objective value will be selected, which increases the computing time. Therefore, how to 1) {\em identify the most effective columns} and 2) {\em decide the appropriate column number} are keys to the performance of FCG. 

In the last few years, researchers have become increasingly interested in  machine learning methods for mathematical optimization including large LPs \cite{BENGIO2021405}. Most related to our work,  \citet{NEURIPS2022_3ecfe5c6} proposed the first Reinforcement Learning (RL) framework for CG, modeling column selection as a Markov Decision Process (MDP). 
Recently, \citet{yuan2023reinforcement} devised a multiple-column selection strategy based on RL for CG. Specifically, they limited the action space size by adding only a \textit{fixed} number of columns per CG iteration. Note that the action space grows exponentially when all possible combinations of candidate columns are considered. Most importantly, when multiple columns are added, the contribution of each column towards convergence varies with the combination of columns in different phases of CG. 
Thus, it is essential to efficiently select the best combination of columns from candidates at each CG iteration to avoid increasing computational overhead with redundant columns. 
This raises the issue of the so-called {\em credit assignment for columns}, which pertains to determining the contribution of each column toward CG convergence.

To address the aforementioned challenges, we propose a novel Fast Family Column Generation, named FFCG. Our main contributions are summarized as follows:

\begin{itemize}
    \item \textbf{Fast multiple-column selection in an iteration:}  We propose a novel RL-based family column generation, which selects a {\em variable} number of columns in an iteration. In each time slot, the size of the action space is reduced from $O(2^n)$ to $O(n^2)$.  We show that FFCG offers better tradeoffs between speeding up the convergence of CG and reducing the total computational time. 
    
    \item \textbf{Reward design and analysis:} 
    To address the credit assignment problem for columns, we carefully design the reward function and evaluate the contribution of each column in each iteration. We also analyze how this design helps FFCG reduce unnecessary computing expenses. 
    
    \item \textbf{Substantial improvements over baselines:} We evaluate FFCG using the common benchmarks for CSP and VRPTW. 
    In our experiments, FFCG converges faster on the benchmarks and reduces the total computing time by 71.4\% and 84.0\% on average compared to several state-of-the-art methods for CSP and VRPTW.
\end{itemize}

The rest of the paper is organized as follows. Firstly, we briefly review previous research on this topic. Subsequently, we present both the standard and RL-based CG methods to solve large LPs. Then, we propose our FFCG framework, which encompasses formulation, analysis, training, and execution. Finally, we assess the effectiveness of FFCG on CSP and VRPTW, in comparison to the baseline approaches, followed by our conclusion and discussion.

\section{Related Work}
In this section, we review the acceleration methods for CG, with a focus on recent advancements in ML-based CG.

\subsection{Acceleration Methods for Column Generation}
To speed up CG, one method is to add multiple columns instead of just one with the most negative reduced cost. \citet{goffin2000multiple} showed that the convergence process can be sped up by selecting non-correlated columns. Then, two practical strategies were proposed for selecting multiple columns~\cite{touati2010solutions}. These strategies were designed to increase the diversity of the selected columns. However, there is no perfect strategy for selecting columns that can minimize the number of iterations for CG while also considering computing costs.

Another approach is dual stabilization, which aims to formulate a better pricing subproblem as the pricing subproblem is a bottleneck for computing time. \citet{doi:10.1287/opre.1060.0278} studied the use of two types of Dual-Optimal Inequalities (DOI) to accelerate and stabilize the whole convergence process, followed by \citet{gschwind2016dual,VACLAVIK20181055,yarkony2020data,haghani2022smooth}. 
We note that our column selection strategy does not conflict with dual stabilization techniques, heuristic, and meta-heuristic methods. They can be used synergistically for further improvement~\cite{YUAN2021102391,pmlr-v235-shen24e}. 

\subsection{Machine-Learning-based Column Generation}

Except for the previously mentioned RLCG \cite{NEURIPS2022_3ecfe5c6}, \citet{mouad-columnselection} proposed machine-learning-based column selection to accelerate CG. The approach applies a learned model to select a subset of columns generated at each iteration of CG. It reduces the computing time spent reoptimizing the RMP at each iteration by selecting the most promising columns.
\citet{babaki2022neural} formulated the task of choosing one of the columns to be included in the RMP as a contextual MDP. They proposed and explored several architectures for improving the convergence of the CG algorithm using deep learning.
However, it also only adds one column at each iteration. Recently, \citet{yuan2023reinforcement} proposed the first RL-based multiple-column selection strategy for CG, which selects $k$ columns from
the pool of $n$ candidate columns generated by the SP. In this approach, $k$ remains fixed. However, selecting more columns in the early stage and fewer columns in the later stage helps to expedite the convergence of CG.

\section{Background}
In this section, we first provide an overview of the standard CG algorithm and then briefly describe the RLCG method.

\subsection{Column Generation in LP}

Let us consider the following generic Linear Program (LP), called the Master Problem (MP):
\begin{equation}
\min \sum\nolimits_{p\in P}c_p\theta_p
\end{equation}
subject to
\begin{equation}
\sum\nolimits_{p\in P} \mathbf{a}_p\theta_p=\mathbf{b}; \ \theta_p\geq 0,\ \forall p\in P
\end{equation}
where $P$ is the index set of variables $\theta_p$; $c_p\in \mathbb{R}$ and $\mathbf{a}_p\in \mathbb{R}^m$ are  the cost coefficient and constraint coefficient vector of $\theta_p$, respectively; and $\mathbf{b}\in \mathbb{R}^m$ is the right-hand-side vector of the constraints. We assume that the number of variables $\theta_p$
is very large and the set of objects associated with these
variables can be implicitly modeled as solutions of an
optimization problem.

The standard CG proceeds to solve this MP as follows.
In each iteration of CG, the RMP that considers a subset $\Omega\subseteq P$ of the columns is solved first. It yields a primal solution $x$ (assuming that $\theta_p=0$, $\forall p\in \Omega\backslash P$) and a dual solution given by the dual values $\pi \in \mathbb{R}^m$ associated with the constraints. Next, the dual solution is used to identify new negative reduced cost variables $\theta_p$, by solving the following pricing subproblem:
\begin{equation}
    \bar{c}=\min_{p\in P}\left\{c_p-\pi^T\mathbf{a}_p\right\}
\end{equation}
If negative reduced cost columns are found, we append them in $\Omega$ to the RMP, and the entire procedure is iterated. Otherwise, CG stops since $\pi$ is an optimal dual solution to the original problem and together with the optimal primal solution to the RMP, i.e., we have an optimal primal/dual pair. 

For some problems, the search for negative reduced cost columns can be distributed across several SPs. When the RMP includes too many columns after several iterations, columns with large reduced cost can be  removed. 

\subsection{Reinforcement Learning for CG (RLCG)}
RLCG \cite{NEURIPS2022_3ecfe5c6} formulates CG as an MDP, denoted as $(\mathcal{S},\mathcal{A},\mathcal{T},r,\gamma)$, where: $\mathcal{S} $ is the state space, $\mathcal{A}$ is the action space, $\mathcal{T}:\mathcal{S}\times\mathcal{S}\times\mathcal{A}\rightarrow[0,1]$, $(s',s, a)\mapsto\mathbb{P}(s'|s,a)$ the transition function, $r:\mathcal{S}\times\mathcal{S}\times\mathcal{A}\rightarrow\mathbb{R}$ the reward function, and $\gamma\in(0,1)$ the discount factor.

At a high level, the method works
as follows. The SP is solved at each iteration and a set of near-optimal column candidates
$\mathcal{G}$ is returned, which is a general feature of optimization solvers such as Gurobi~\cite{gurobi}. RLCG selects a column from $\mathcal{G}$ according to the Q-function learned by the RL agent. The RL agent is fused
within the CG loop and actively selects the column to be added to the next iteration using the information extracted
from the current RMP and SP.

\begin{figure*}
    \centering
    \includegraphics[width=0.97\textwidth]{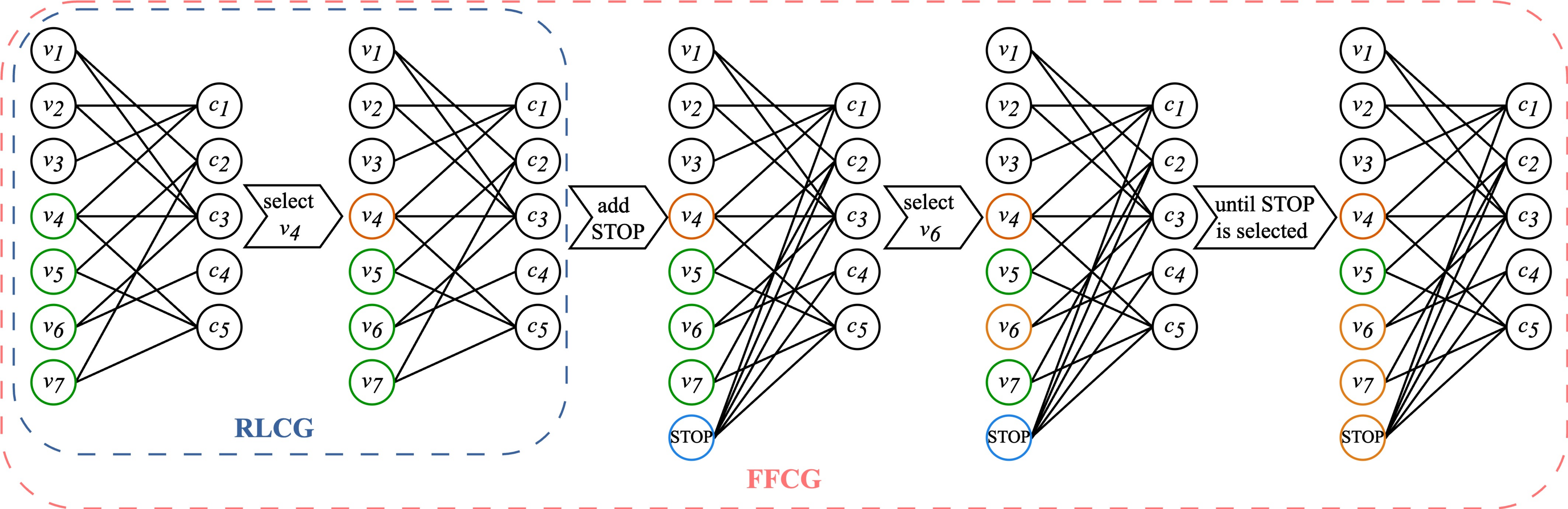}
    \caption{Column Selection in FFCG: First, the available action set is $\mathcal{G}=\{v_4,v_5,v_6,v_7\}$ (green node), and the RL agent selects $v_4$ (the selected node is orange). Then, a new available action $\mathrm{STOP}$ (blue node), which means stop column selection and return the selected columns, is added to action space $\mathcal{G}-\{v_4\}$. The RL agent repeatedly updates the state and selects from the remaining actions until the action $\mathrm{STOP}$ is selected. Finally, the suggested columns set $\mathcal{C}=\{v_4,v_6,v_7\}$ is returned.}
    \label{fig:ffcg}
\end{figure*}

The state $S$ is represented by the bipartite graph of the current CG iteration from the RMP and the candidate columns from the SP.  As shown in the left part of Figure \ref{fig:ffcg}, the RMPs are encoded using bipartite graphs with columns (variables) nodes ($v$) and constraint nodes ($c$) \cite{10.5555/3454287.3455683}.  The edge between $v$ and $c$ in the graph indicates the contribution of a column $v$ to a constraint $c$. Note that each node is characterized by its feature vector. Each action in the action set $\mathcal{G}$ represents a candidate column (green node). The RL agent selects one column (action) $a$ to add to the RMP for the next iteration from the candidate columns set $\mathcal{G}$ returned from the current iteration SP. Transitions are deterministic. After selecting an action from the current candidate set $\mathcal{G}$, the selected column enters the basis in the next RMP iteration. The detailed definition of the reward function is given by \citet{NEURIPS2022_3ecfe5c6}. A GNN is used as a Q-function approximator, trained with experience replay \cite{mnih2015human}. 

\section{FFCG Framework}

Here, we propose our FFCG framework.
At a high level, FFCG works very similar to RLCG. At each iteration, we solve the RMP to obtain dual values and use the dual values to update the SP objective function. Then, the SP is solved. We define a solution gap for the worst-case candidate column, ensuring that its reduced cost remains close to the optimal value, thus making it a near-optimal candidate for selection. Subsequently, the set of candidate columns $\mathcal{G}$ is returned. With $\mathcal{G}$ and the context $S$ in the bipartite graph, it selects a set of columns $\mathcal{C}$ to be added to the next iteration. This is the key difference from RLCG where only a single column is selected. While no column with negative reduced cost exists, it stops and the optimal solution is returned.
Now, there are two key challenges: 1) {\em how to select effective columns from the candidates} and 2) {\em when to stop column selection}.
 %


 \subsection{MDP Formulation}
To formulate FCG as an MDP, a candidate column $a_{i,t}$ in each CG iteration is called an available action. Let the candidate columns in the $t^{\mathrm{th}}$ CG iteration $\mathcal{G}_t = \{a_{1,t}, \dots , a_{|\mathcal{G}_t|,t}\}$ be the set of available actions in the time slot $t$. Note that this formulation captures the volatile actions: action sets $\mathcal{G}_t$, $\forall ~ 0<t\leq T$ (and their size) in different time slots can be different from each other. For each available action $a_{i,t}$, its context (side information) can be observed. Let $R_t(\mathcal{C}_t)$ be the quality of the selected column set $\mathcal{C}_t$ (the reward of choosing $\mathcal{C}_t$ based on the observed context $S_t$ in time slot $t$). 
Given the available actions $\mathcal{G}_t$ to choose from in each time slot, our objective is to pick a subset of actions $\mathcal{C}_t \subseteq \mathcal{G}_t$  to maximize the total reward.

At each iteration (time slot), SP returns a set of columns $\mathcal{G}_t$ with negative reduced costs. Then every possible combination of candidate columns can be returned as $\mathcal{C}_t$. In other words, in each iteration, the RL agent can select one or more columns to be added to the next iteration. For example, as shown in the left-hand side of Figure \ref{fig:ffcg}, the set of available columns (the green column nodes) is $\mathcal{G}_t=\{v_4,v_5,v_6,v_7\}$, and the set of available $\mathcal{C}_t$ is $\mathcal{P}(\mathcal{G}_t)\backslash\{\emptyset\}$, where $\mathcal{P}(\mathcal{G}_t)$ is the power set of $\mathcal{G}_t$. Although adding more columns can reduce the total number of iterations, it also increases the computing costs. Therefore, the RL agent should learn to select the most effective columns set $\mathcal{C}^*_t$, balancing both speed and costs. 

In what follows, we will describe how to select a subset of columns $\mathcal{C}_t \subseteq \mathcal{G}_t$ based on the context $S$.

\subsection{Column Selection}

To avoid the exponential growth in the action space, we select the candidate columns one by one. In other words, we turn the factored action space into a sequential choice, inspired by \citet{NEURIPS2022_69413f87}, and only one column is selected in each turn. Note that this is similar to the local search in the huge action space, where a move in each dimension is decided at a time. Although it has no guarantee to select the optimal combination, this simple strategy performs well in practice given a good marginal value, as shown later in our experiments. Moreover, it significantly reduces the action space considered in RL. There are $O(|\mathcal{G}_t|)$ actions in each step, and the number of steps in each time slot is $O(|\mathcal{G}_t|)$. Thus, the action space size of FFCG is $O(|\mathcal{G}_t|^2)$, while the action space size of RLCG is $O(2^{|\mathcal{G}_t|})$. In addition, we add the STOP action (a blank column) to the action space for convenience. When the STOP action is selected, the column selection is stopped and the selected columns are returned. 

\begin{algorithm}[tb]
\caption{Column Selection}
\label{alg:agent}

\textbf{Input}: Actions set $\mathcal{G}$, and context $S$\\
\textbf{Output}: Suggested columns set $\mathcal{C}$
\begin{algorithmic}[1] 
\STATE $\mathcal{C}\leftarrow\emptyset$, $k\leftarrow0$
\STATE $\hat{Q}_{\Delta}\leftarrow\mathrm{MarginalQValueApproximator}(\mathcal{G},\mathcal{C},S)$
\STATE $a_k\leftarrow \mathop{\arg\max}\nolimits_{a\in \mathcal{G}}\hat{Q}_{\Delta}(\mathcal{C}\cup\{a\}, a)$
\STATE $\mathcal{C}\leftarrow\mathcal{C}\cup\{a_k\}$ \hfill$\triangleright$ \textit{Select a column}
\STATE $\mathcal{G}\leftarrow(\mathcal{G}-\{a_k\})\cup\{\mathrm{STOP}\}$ \hfill$\triangleright$ \textit{Add STOP action}
\STATE $S\leftarrow\mathrm{UpdateContext}(\mathcal{G},\mathcal{C},S)$
\STATE $k\leftarrow k+1$
\WHILE{STOP is not selected}
\STATE $\hat{Q}_{\Delta}\leftarrow\mathrm{MarginalQValueApproximator}(\mathcal{G},\mathcal{C},S)$
\STATE $a_k\leftarrow \mathop{\arg\max}\nolimits_{a\in \mathcal{G}}\hat{Q}_{\Delta}(\mathcal{C}\cup\{a\}, a)$
\STATE $\mathcal{C}\leftarrow\mathcal{C}\cup\{a_k\}$ \hfill$\triangleright$ \textit{Select a column}
\STATE $\mathcal{G}\leftarrow\mathcal{G}-\{a_k\}$
\STATE $S\leftarrow\mathrm{UpdateContext}(\mathcal{G},\mathcal{C},S)$
\STATE $k\leftarrow k+1$

\ENDWHILE
\STATE \textbf{return} $\mathcal{C}\leftarrow\mathcal{C}-\{\mathrm{STOP}\}$

\end{algorithmic}
\end{algorithm}

Algorithm \ref{alg:agent} outlines the main procedure utilized by the RL agent to make its selection of suggested columns. Firstly, the agent gets the expected marginal Q-value $\hat{Q}_{\Delta,t}$ of each action (select a candidate column). The action with the greatest expected marginal Q-value would be added greedily in the suggested columns set $\mathcal{C}_t$. Then, the STOP action is added to the action space $\mathcal{G}_t$, and the last selected action is removed. We repeat the above steps until the 
 STOP action selected. In the end, the STOP action (a blank column) is removed from $\mathcal{C}_t$ and the suggested columns set is returned. 

To learn the marginal Q-value, the key challenge is how to approximately compute the expected marginal reward.

\subsection{Reward Design}

For given suggested columns $\mathcal{C}_t$, we design the reward function consisting of two components: 1) the change in the RMP objective value,
where a bigger decrease in value is preferred; and 2) the penalty for redundant columns, which are no improvements to the objective value. Together, they incentivize the RL agent to converge faster and avoid selecting redundant columns. 

Specifically, the reward of a subset of columns $\mathcal{C}_t$ at time slot $t$ is defined as:
\begin{equation}\label{eq4}
R_t(\mathcal{C}_t)\triangleq\alpha\left(\frac{obj_{t-1}-obj_t}{obj_0}\right)-\beta|\mathcal{C}_t-\mathcal{C}'_t|
\end{equation}
where $obj_0$ is the objective value of the RMP in the first CG iteration, which is used to normalize the objective value change ($obj_{t-1} -obj_t$) across instances of various sizes; $\alpha$ is a non-negative hyperparameter that weighs the normalized objective value change in the reward; $\mathcal{C}'_t\subseteq \mathcal{C}_t$ is the optimal subset of $\mathcal{C}_t$, in which all columns make improvement to the RMP objective value; $|\mathcal{C}_t-\mathcal{C}'_t|$ counts the number of redundant columns that are in $\mathcal{C}_t$ but not in $\mathcal{C}'_t$; $\beta$ is a non-negative hyperparameter that weighs the penalty of selecting redundant columns. 
Increasing the values of parameters $\alpha$  will enable the RL agent to choose more columns. On the other hand, a higher value of $\beta$ will prevent the RL agent from selecting too many unnecessary columns. 

\subsubsection{Marginal rewards.} 
Note that the aforementioned reward is the total reward of the suggested columns $\mathcal{C}_t$. However, the total rewards achieved by selected columns are not a simple sum of individual rewards but demonstrate a feature of diminishing returns determined by the relations between selected columns (e.g. relevance and redundancy) \cite{Submodular}. In Algorithm \ref{alg:agent}, we only use the marginal Q-value of each column instead of the total Q-value to form $\mathcal{C}_t$. To this end, we must assign the credit to each column and compute the marginal reward of an individual column.

Here, we denote the marginal reward of an individual column $a$ $(a\in\mathcal{C}_t)$  to a set $\mathcal{C}_t$ by 
\begin{equation}
    r_{\Delta,t}(\mathcal{C}_t,a)\triangleq R_t(\mathcal{C}_t)-R_t(\mathcal{C}_t-\{a\})
\end{equation}
For redundant columns that do not improve the CG convergence ($a\notin \mathcal{C}'_t$), 
the marginal reward of them is $-\beta$. For effective columns ($a\in \mathcal{C}'_t$), the contribution weight of each column $a$ is
\begin{equation}
    \phi_t(\mathcal{C}_t,a)\triangleq\frac{r_{\Delta,t}(\mathcal{C}_t,a)}{\sum_{a'\in \mathcal{C}'_t} r_{\Delta,t}(\mathcal{C}_t,a')}
\end{equation}

Now, the reward of each column is proportional to its marginal reward. Thus, the reward of an individual column $a\ (a\in\mathcal{C}_t)$ in time slot $t$ is
\begin{equation}
        r_{\Delta,t}(\mathcal{C}_t,a)=\left\{\begin{array}{lr}
\phi_t(\mathcal{C}_t,a) R_t^{obj}(\mathcal{C}_t) & a\in \mathcal{C}'_t, \\
        -\beta & a\notin \mathcal{C}'_t.
    \end{array}\right.
\end{equation}
where the term $R_t^{obj}(\mathcal{C}_t) = \alpha (\frac{obj_{t-1}-obj_t}{obj_0})$ 
denotes the total contribution of all effective columns $a\in \mathcal{C}'_t$ to the objective value change, i.e., $R_t(\mathcal{C}_t)$ without the second term.

For the STOP action, the reward is 0 because it does not improve the convergence of CG and does not increase unnecessary computing expenses. For the unselected action $a\in \mathcal{G}_t-\mathcal{C}_t$, if it should be selected ($r_{\Delta,t}(\mathcal{C}_t\cup\{a\},a)>0$), then the reward is $\beta$, otherwise the reward is $-\beta$. 

\begin{algorithm}[!t]
\caption{FFCG Training and Execution}
    \label{alg1}
    \textbf{Input}: Problems $\{p_i\}_{i=1}^{M}$ and hyperparameters $\alpha, \beta$ \\
    \textbf{Output}: Q function approximator $\hat{Q}^*$ at training time or optimal solutions at execution time
    
    \begin{algorithmic}[1]
    \IF{at training time}
    \STATE Initialize replay memory $D$ to capacity $N$
    \STATE Initialize Q function approximator $\hat{Q}$ with random weights $\theta$ and target $\hat{Q}^*$ with weights $\theta^-=\theta$ 
    \ENDIF
    \FOR{$i\leftarrow1$ to $M$}
    \STATE $t\leftarrow0$
    \STATE $\mathrm{RMP}_0\leftarrow \mathrm{Initialize}(p_i)$
    \STATE Solve $\mathrm{RMP}_0$ to get dual values
    \STATE Use dual values to construct $\mathrm{SP}_0$
    \STATE $\mathcal{G}_0\leftarrow \mathrm{GetCandidateColumns}(\mathrm{RMP}_0,\mathrm{SP}_0)$
    
    \WHILE{CG algorithm has not converged $(\mathcal{G}_t\neq \emptyset)$}
    \STATE $S_t\leftarrow \mathrm{Context}(\mathrm{RMP}_t,\mathrm{SP}_t,\mathcal{G}_t)$
    \STATE $\mathcal{C}_t\leftarrow \mathrm{ColumnSelection}(\mathcal{G}_t,S_t)$ \hfill$\triangleright$ \textit{Select columns}
    \IF{at training time and with probability $\epsilon$}
    \STATE Select a random subset $\mathcal{C}_t$ from $\mathcal{G}_t$ instead
    \ENDIF
    \STATE Add columns in $\mathcal{C}_t$ to $\mathrm{RMP}_t$  and get $\mathrm{RMP}_{t+1}$
    \IF{at training time}
    \STATE Observe reward $R_t(\mathcal{C}_t)$ and $r_{\Delta,t}(\mathcal{C}_t,a)$  $\forall a\in\mathcal{C}_t$
    \STATE Store transition $(RMP_t,\mathcal{C}_t,r_t,RMP_{t+1})$ in   $D$
    \STATE $(RMP_j,\mathcal{C}_j,r_j,RMP_{j+1}) \gets$ Sample random minibatch of transitions from $D$
    \STATE Perform a gradient descent step w.r.t network parameters $\theta$ on $[Q_{\Delta,t}(\mathcal{C}_t, a)-\hat{Q}^*_{\Delta,t}(\mathcal{C}_t, a)]^2$
    \STATE Reset $\hat{Q}^*\leftarrow\hat{Q}$ in every $C$ steps 
    \ENDIF
    \STATE Solve $\mathrm{RMP}_{t+1}$ to get dual values
    \STATE Use dual values to build $\mathrm{SP}_{t+1}$
    \STATE $t\leftarrow t+1$
    \STATE $\mathcal{G}_t\leftarrow \mathrm{GetCandidateColumns}(\mathrm{RMP}_t,\mathrm{SP}_t)$
    \ENDWHILE
    \IF{at execution time}
    \STATE \textbf{return} Optimal solutions from $\mathrm{RMP}_t$, $\mathrm{SP}_t$
    \ENDIF
    \ENDFOR
    \IF{at training time}
    \STATE \textbf{return} Trained Q function approximator  $\hat{Q}^*$
    \ENDIF
\end{algorithmic}
\end{algorithm}

\subsection{Training and Execution}

Theoretically, given the total reward $R_t(\mathcal{C}_t)$, we can enumerate all possible $\mathcal{C}_t \subseteq \mathcal{G}_t$ and select the best one. However, this brute-force method is not scalable because the number of $\mathcal{C}_t$ grows exponentially with the number of candidate columns. Therefore, we compute the marginal Q-value of an individual column $Q_{\Delta,t}$ based on $R_t(\mathcal{C}_t)$, and select the suggested columns as shown in Algorithm \ref{alg:agent}. This is more practical in terms of the computational costs. Unfortunately, the total reward $R_t(\mathcal{C}_t)$ depends on $obj_t$ and $\mathcal{C}'_t$ both of which are unknown before column selection. Therefore, we train a marginal Q-value  approximator $\hat{Q}$ for $Q_{\Delta,t}$ with experience replay \cite{mnih2015human}. This is done by minimizing the mean squared loss between $Q_{\Delta,t}(\mathcal{C}_t,a)$ and $\hat{Q}_{\Delta,t}(\mathcal{C}_t,a)$. The FFCG training and execution is shown in Algorithm \ref{alg1}.

Specifically, at the training time, we use a $\epsilon$-greedy strategy to select columns and compute $r_{\Delta,t}(\mathcal{C}_t,a)$ based on $R_t(\mathcal{C}_t)$ after observing $obj_t$ and $\mathcal{C}'_t$. Then, we perform a gradient descent step to update the network parameters $\theta$ of the marginal Q-value approximator. During execution, we use the RL agent with the marginal Q-value approximator to select columns according to Algorithm \ref{alg:agent}. After that, the CG proceeds to the next iteration until it converges. 

Note that the key difference between Algorithm \ref{alg1} and RLCG is the column selection policy. It is replaced by our polynomial-time multiple-column selection algorithm (Algorithm~\ref{alg:agent}) to accelerate CG. The way we train the marginal Q-value approximator is similar to DQN \cite{mnih2015human} that RLCG uses to learn the Q-function. To some extent, RLCG can be viewed as a special case of our method, when we only select one column in Algorithm \ref{alg:agent}. Compared to RLCG, the major improvements in our approach are the total and marginal reward design and the selection strategy for multiple columns. 

\section{Experiments}
We evaluate our proposed FFCG on two sets of problems: the CSP and the VRPTW.
Both problems are well-known for the linear relaxation effectively solved using CG. Experimental results demonstrate
that FFCG outperforms several widely used
single-column selection strategies and multiple-column selection strategies. Furthermore, we  analyze how FFCG speeds up CG convergence.

\subsection{Experimental Tasks}

\paragraph{Cutting Stock Problem.}
The CSP revolves around efficiently dividing standard-sized stock materials, like paper rolls or sheet metal, into specified sizes while minimizing the surplus material that goes to waste. Computationally, it delves into NP-hard territory and can be reduced to the knapsack problem. Given the combinatorial intricacy inherent in the CSP and its formidable array of potential patterns (variables), the CG technique emerges as a pragmatic solution. It adeptly addresses the LP relaxation of the CSP through an iterative approach, bypassing the exhaustive enumeration of all feasible patterns. The detailed formulation is given in Appendix A. 

\begin{table*}[tb]
\begin{center}
\begin{tabular}{c|rrrrrrrrrrrr}
\toprule
\multirow{2}{*}{Strategy} & \multicolumn{2}{c|}{$n=50$}  & \multicolumn{2}{c|}{$n=200$}  & \multicolumn{2}{c|}{$n=750$} & \multicolumn{2}{c}{$n=1000$}  \\ 
 & \multicolumn{1}{c}{\#Itr} & \multicolumn{1}{c|}{Time} & \multicolumn{1}{c}{\#Itr} & \multicolumn{1}{c|}{Time} & \multicolumn{1}{c}{\#Itr} & \multicolumn{1}{c|}{Time} & \multicolumn{1}{c}{\#Itr} & \multicolumn{1}{c}{Time} \\ \midrule
Greedy-S & 53.10& \multicolumn{1}{r|}{4.19} & 147.30& \multicolumn{1}{r|}{9.47}  & 222.20& \multicolumn{1}{r|}{16.96}  &  386.14&   67.34\\
MLCG-S &  43.48& \multicolumn{1}{r|}{5.10}& 145.74& \multicolumn{1}{r|}{21.78} & 232.93& \multicolumn{1}{r|}{31.93}  &   295.09 &   70.89\\
RLCG-S &  43.21& \multicolumn{1}{r|}{3.76}&  152.80& \multicolumn{1}{r|}{8.77}  & 237.67 & \multicolumn{1}{r|}{16.10}  & 300.86   &   46.88\\\midrule
Greedy-M &  \textbf{11.80}&  \multicolumn{1}{r|}{1.34}&  \textbf{35.15}& \multicolumn{1}{r|}{2.69}  &\textbf{51.13} & \multicolumn{1}{r|}{5.38}  & \textbf{ 66.45}  & 11.51\\
RLMCS-M &  13.87&  \multicolumn{1}{r|}{1.69}&  45.67& \multicolumn{1}{r|}{5.59} &   71.33&\multicolumn{1}{r|}{12.93} &  86.50   &17.80\\
FFCG (Ours) &  11.81& \multicolumn{1}{r|}{\textbf{1.30}} &  38.57& \multicolumn{1}{r|}{\textbf{2.51}} &56.27  & \multicolumn{1}{r|}{\textbf{4.50}}  & 78.86  &  \textbf{9.65}\\ \bottomrule
\end{tabular}
\caption{Experimental results on CSP with different size $n$. It reports the average number of iterations per instance, and the total runtime in seconds (lower is better). }\label{tab:csp}
\end{center}
\end{table*}
\paragraph{Vehicle Routing Problem with Time Windows.} The Vehicle Routing Problem (VRP)  involves finding a set of minimum-cost vehicle routes, originating and terminating at a central depot, that together cover a set of customers with known demands.  Each customer has a given demand and is serviced exactly once, and all the customers must be assigned to vehicles without exceeding vehicle capacities. VRPTW is a variation of the VRP where the service at any customer must be started within a given time interval, called a time window. The detailed formulation of VRPTW is given in Appendix B. We use the well-known Solomon benchmark \cite{Solomon1987AlgorithmsFT} for training and testing. The dataset generation and division are described in Appendix F.
\subsection{Hyperparameter Configuration}
We meticulously fine-tune the central hyperparameters, $\alpha$ and $\beta$, embedded within the reward function (Equation \ref{eq4}). The process encompasses an exhaustive grid spanning 25 potential configurations, from which we sample 14 distinct setups. These configurations are subsequently employed to train the agent, and their efficacy is gauged using a dedicated validation set. Further elaboration on this methodology is available in Appendix E.
The specific parameter configuration adopted for addressing both CSP and VRPTW is as follows: $\alpha=2000$, $\beta=0.3$. 

\subsection{Comparison Evaluation}
We compare our FFCG with several well-established single-column and multiple-column selection strategies. We select the same number of candidate columns for all column selection strategies. The candidate columns are generated as the 10 columns with the most negative reduced cost from SP. The baseline strategies for comparison are as follows.
\paragraph{Single-column selection strategies:}
\begin{itemize}
    \item {\bf Greedy-S}: 
    the traditional greedy strategy that selects the column with the most negative reduced cost at each step.
    \item {\bf MLCG-S}: selection strategy using the learned
MILP expert in \cite{mouad-columnselection}.
    \item {\bf RLCG-S}: the RL-based single-column selection strategy by \citet{NEURIPS2022_3ecfe5c6}.
\end{itemize}

\paragraph{Multiple-column selection strategies:}
\begin{itemize}
    \item {\bf Greedy-M}: the simple multiple-column selection strategy that selects all candidate columns with the negative reduced costs at each step.
    \item {\bf RLMCS-M}: the RL-based multiple-column selection strategy which selects 5 columns per iteration~\cite{yuan2023reinforcement}.
\end{itemize}

 We measure 1) the average number of iterations per instance, 2) the average number of columns added per instance, and 3) the total runtime in seconds, which includes GNN inference and feature computations (if applicable). The node features of GNN are described in Appendix D. For the single-column selections, the number of selected columns is equal to the number of iterations. Thus, we only compare this term for multiple-column selections.
\subsection{Experimental Results}

\begin{table}[tb]
    \centering
    \begin{tabular}{c|c|c|c|c}\toprule
         Strategy&  $n=50$& $n=200$ & $n=750$ &$n=1000$ \\\midrule
        Greedy &  117.97&  351.52&  511.33& 664.55\\
        RLMCS &  69.37&  228.37&  356.67& 432.50\\
        FFCG&  77.99&  250.63&  377.40& 462.73\\\bottomrule
    \end{tabular}
 \caption{Experimental results on CSP with different size $n$. It reports the average number of columns added, which is only compared between multiple-column selection strategies.}
    \label{tab:my_label}
\end{table}
\paragraph{Results on CSP.}
 We first train FFCG on CSP instances with the roll length $n=50,100,200$ and the number of item types $m$ varying from $50$ to $150$. Employing a curriculum learning strategy \cite{10.5555/3455716.3455897}, FFCG is trained by feeding the instances in order of increasing difficulty. We test FFCG and other compared methods using CSP instances with the roll length $n=50,200,750,1000$. The detailed dataset information is described in Appendix F.

\begin{figure*}[tb]%
    \centering
        \subfloat[$n=50$, test instances]{
        \includegraphics[width=0.24\linewidth]{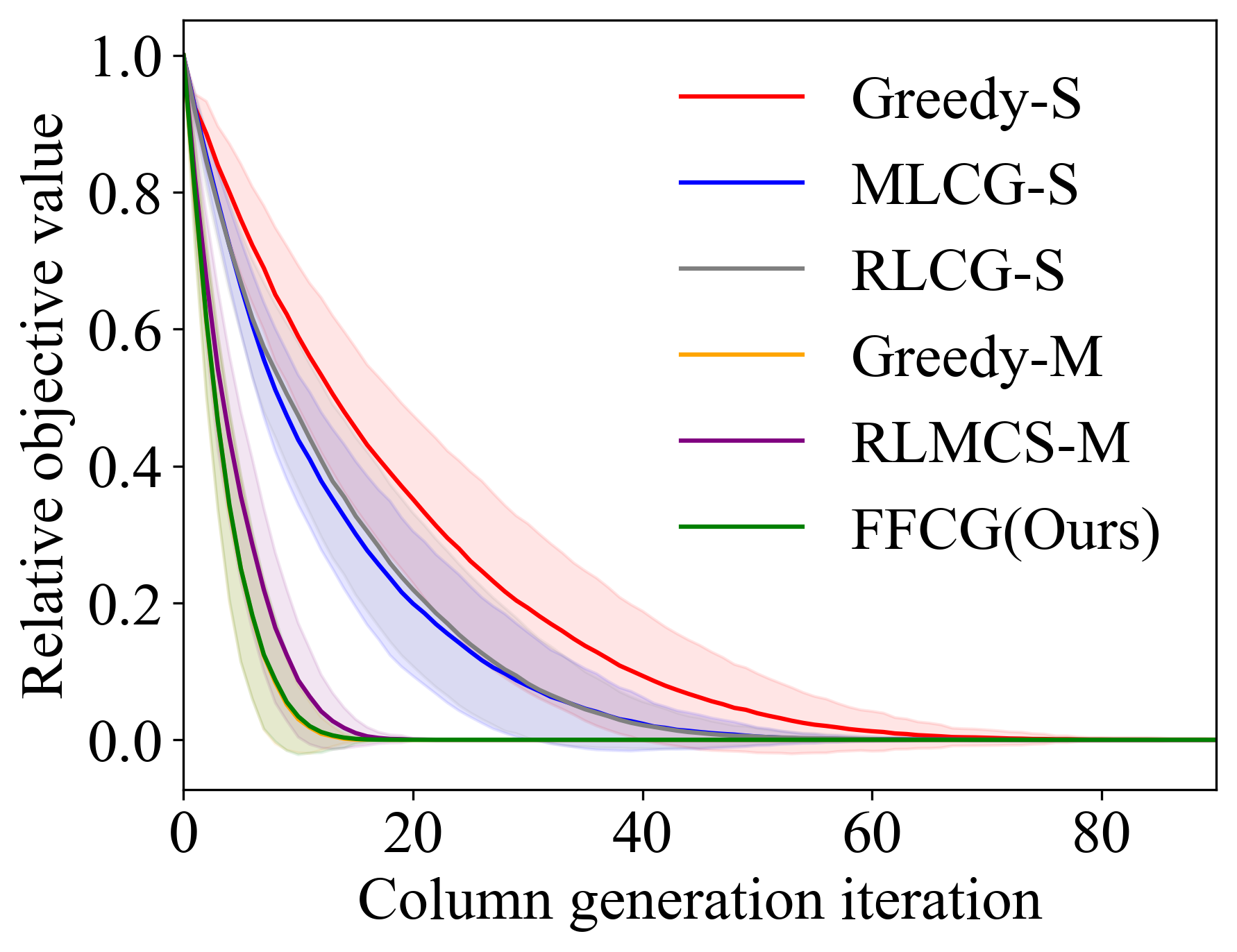}
        }\hfill
        \subfloat[$n=200$, test instances]{
        \includegraphics[width=0.24\linewidth]{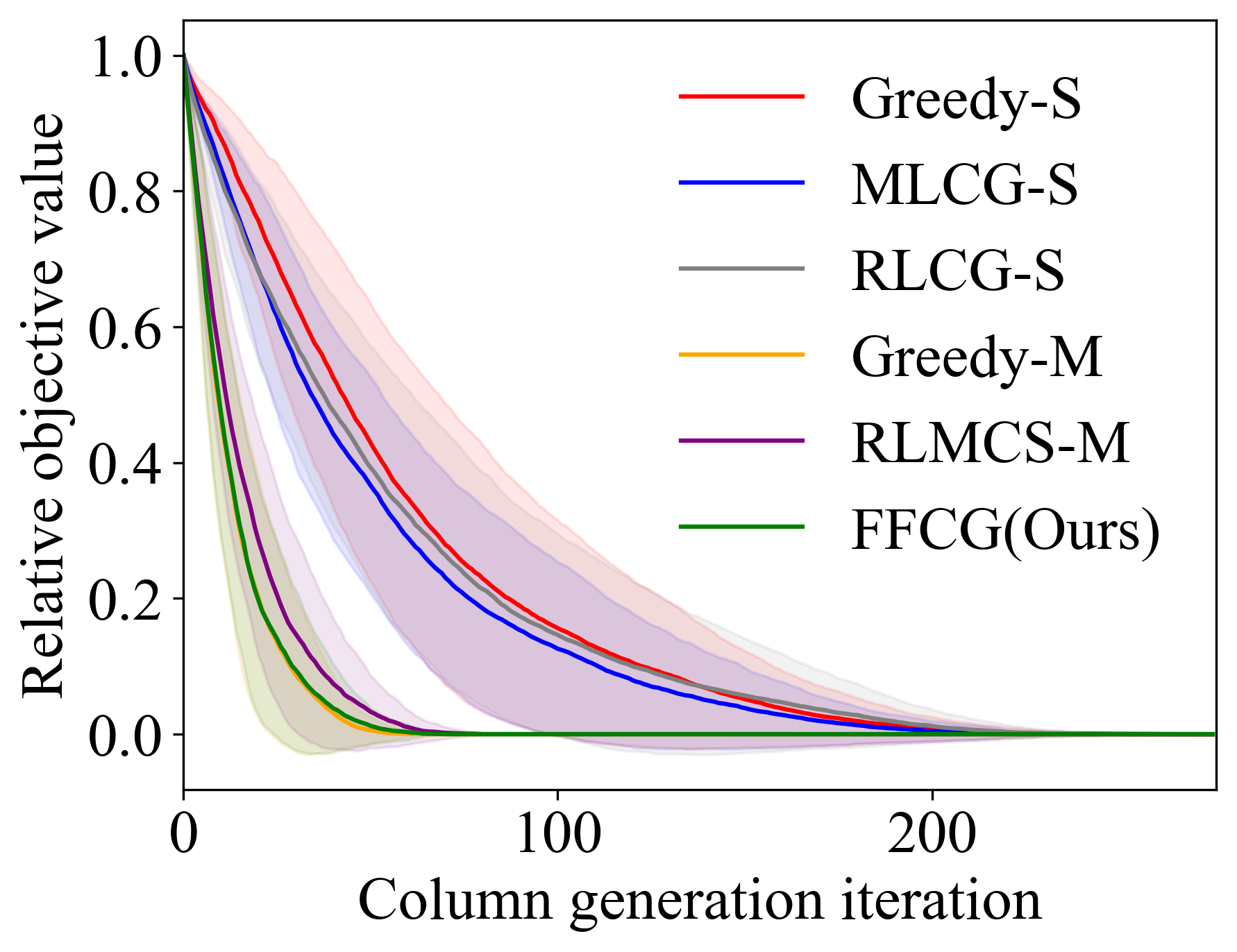}
        }\hfill
    \subfloat[$n=750$, test instances]{
        \includegraphics[width=0.24\linewidth]{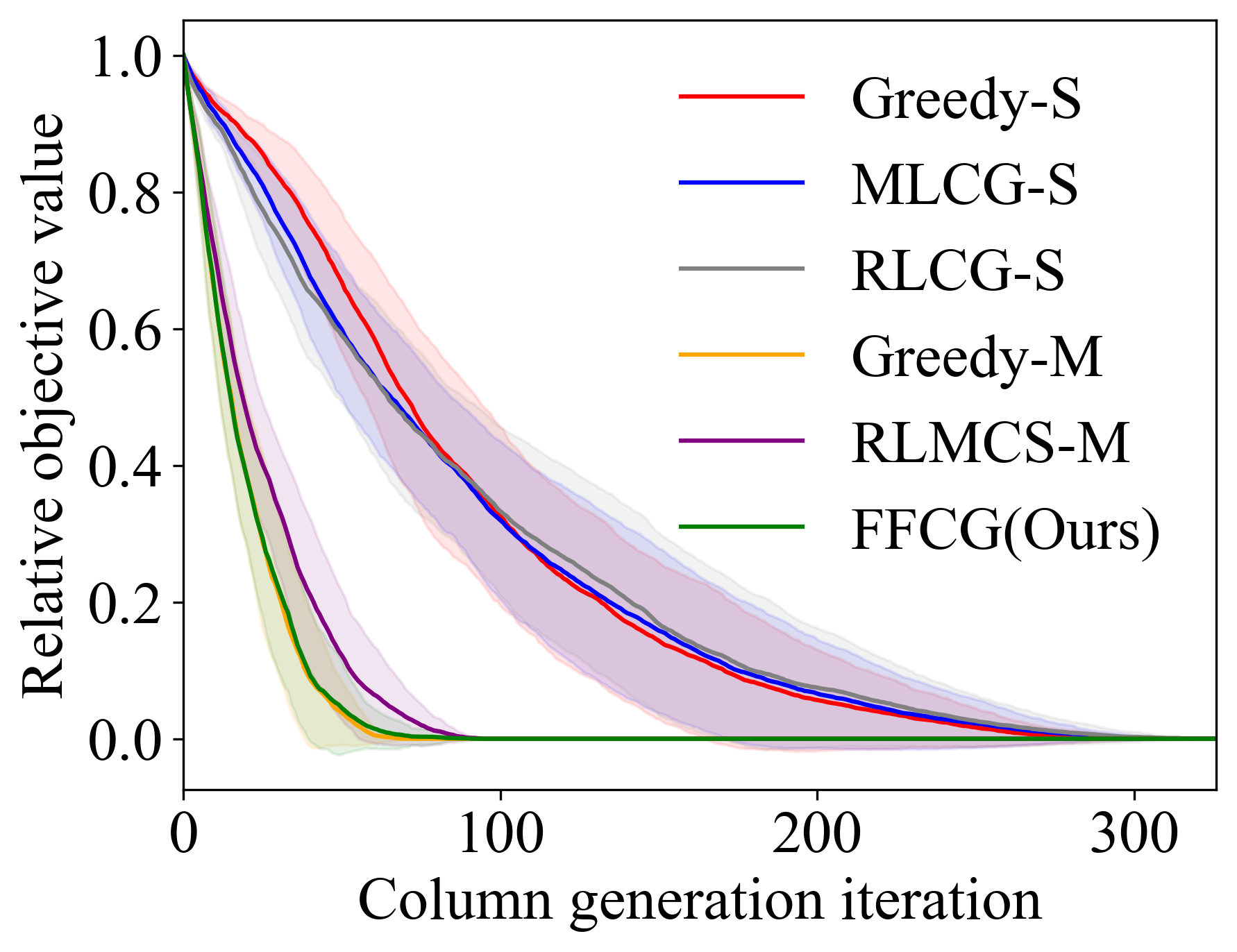}
        }
    \hfill
    \subfloat[$n=1000$, test instances]{
        \includegraphics[width=0.24\linewidth]{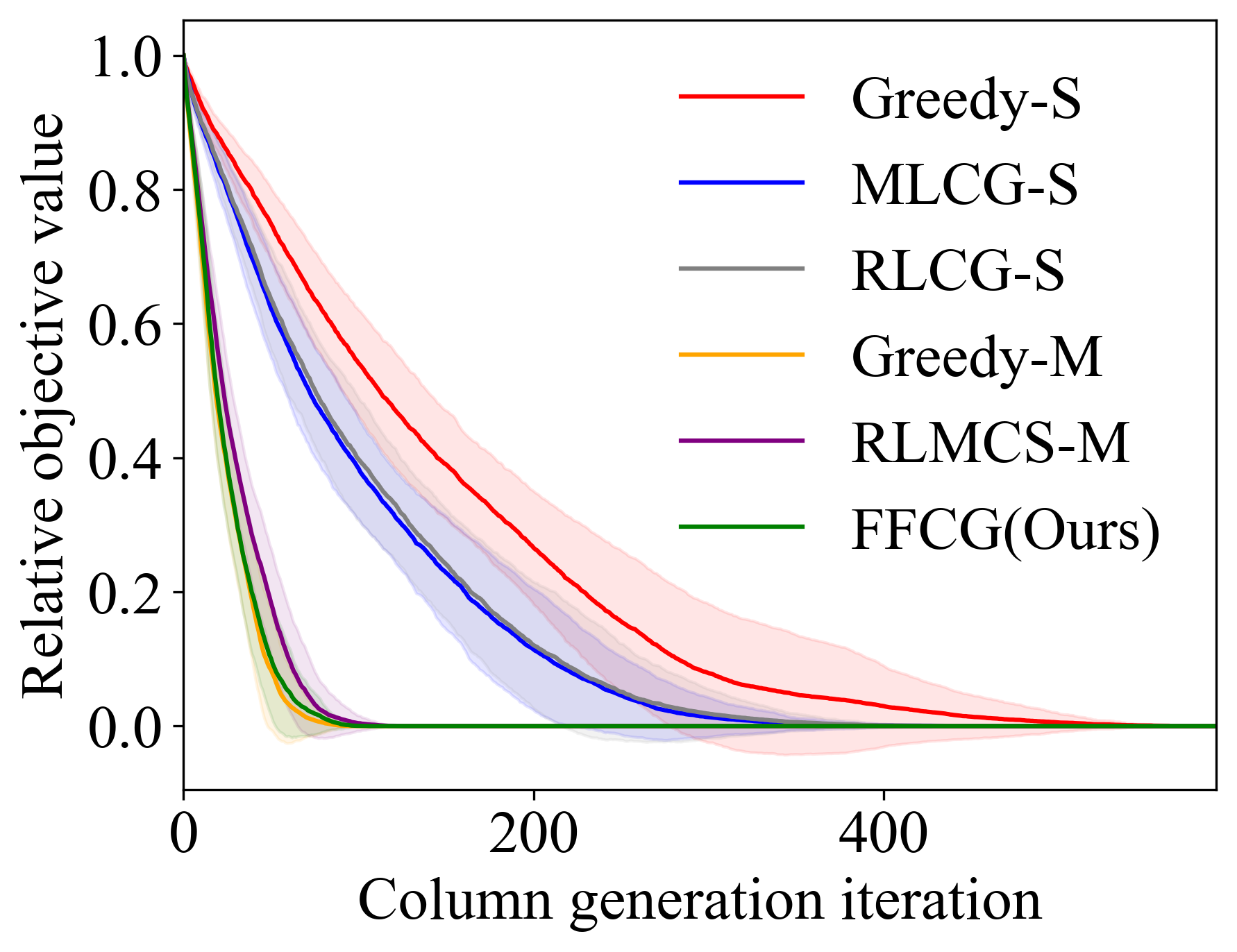}
        }\\      
    \caption{CSP: CG convergence plots for FFCG (green), Greedy-S (red), MLCG-S (blue), RLCG-S (gray), Greedy-M (orange), and RLMCS-M (purple). The solid curves are the mean of the objective values for all instances, and the shaded area shows the standard deviation.}
    \label{4figs}
\end{figure*}
The comparison results on CSP are reported in Table \ref{tab:csp}, Table \ref{tab:my_label}, and Figure \ref{4figs}. To illuminate the comparative performance statistically,  we visualize the CG-solving trajectories for all test instances with different roll lengths $n$. We record the objective values of the RMP at each CG
iteration for given methods, then take the average over all instances. Note that we normalized the objective values
to be in $[0,1]$ before taking the average among instances. The result shows that FFCG outperforms the prior state-of-the-art methods. 
Remarkably, despite not being directly trained on CSP instances with roll lengths of 750 and 1000, FFCG demonstrates commendable performance in such challenging scenarios, thereby reflecting its robust generalization capabilities.
Compared with Greedy-S, FFCG reduces the number of CG iterations by 77.1\% and reduces the total computing time by 71.4\% for all test instances on average. 

As shown, FFCG performs better than Greedy-M on harder problems. Although selecting all candidate columns reduces the number of iterations, it ends up selecting too many redundant columns and increases the computing cost. 
\paragraph{Results on VRPTW.}

\begin{table}[tb]
\begin{center}
\begin{tabular}{c|rrr} \toprule
 Strategy& \multicolumn{1}{c}{\#Itr} & \multicolumn{1}{c}{Time} &    \multicolumn{1}{c}{\#Col}\\ \midrule
Greedy-S &  28.50& \multicolumn{1}{r}{2662.07} &    28.50\\
RLCG-S &  16.50& \multicolumn{1}{r}{1789.98} &    16.50\\\midrule
Greedy-M &  4.00& \multicolumn{1}{r}{511.97} &    40.00\\
FFCG (Ours) &  4.33& \multicolumn{1}{r}{426.99} &    24.33\\ \bottomrule
\end{tabular}
\caption{Experimental results on VRPTW. It reports the average number of iterations per instance, the total runtime in seconds, and the average number of columns added.}
    \label{tab:vrp}
\end{center}
\end{table}

The results are shown in Table \ref{tab:vrp}. Compared with the Greedy-S method, FFCG reduces the number of CG iterations by 84.8\%  and the total computing time by 84.0\%.  Notice that we do not compare FFCG with MLCG and RLMCS because they do not design features for VRPTW. The convergence plot is given in Appendix H.

\subsection{Analysis}

\begin{figure}[tb]
    \centering
    \includegraphics[width=0.45\textwidth]{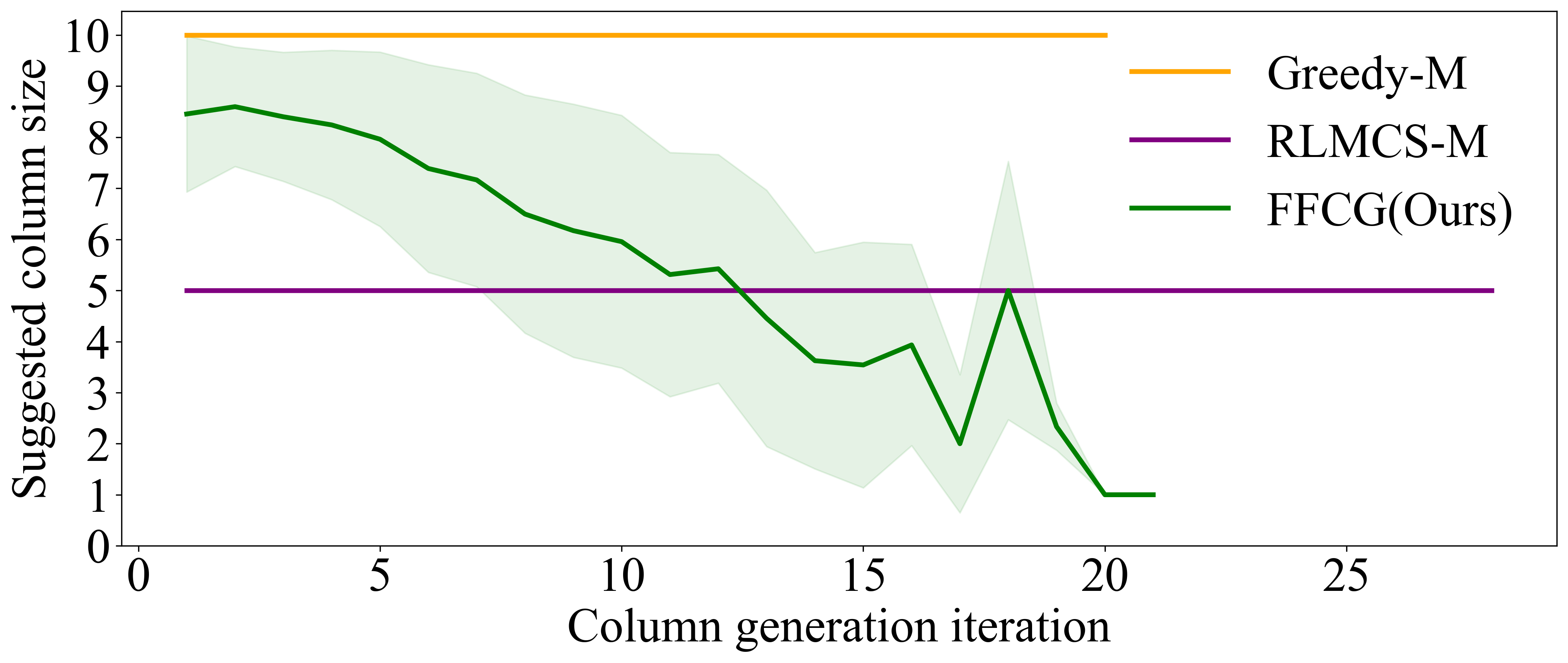}
    \caption{CSP, $n=50$ test instances: The suggested column size per CG iteration for FFCG, Greedy-M, and RLMCS-M.}
    \label{fig:Analysis}
\end{figure}

We further discuss why FFCG converges faster than other strategies by examining the distribution of the number of columns selected in each iteration. The cases of selecting all or only one column represent only 30.50\% of the total. In major cases, the selection dynamically adjusts between these two extremes. As shown in Figure \ref{fig:Analysis}, in the initial rounds, FFCG selects a greater number of columns that contribute to improving the objective function (contributory columns) to accelerate the convergence of CG. 
The number of contributory columns in the candidate list decreases as the CG processes. 
At this stage, FFCG adjusts the number of selected columns to avoid selecting redundant columns, which would bring additional computational overhead. 
As the saying goes, to improve is to change, to be perfect is to change often.
In this context, fixing the number of columns selected in each round, as done in RLMCS, is not an effective strategy. In contrast, FFCG dynamically adjusts the number of suggested columns in each iteration, based on the quality of the candidate columns and the columns in the MP.

\section{Conclusion and Discussion}

In this paper, we propose FFCG, a novel RL-based multiple-column selection framework for CG. Specifically, we model the column selection as an MDP problem. Then, we devise a $O(n^2)$ algorithm to select a subset of candidate columns based on the observed context. 
Moreover, we learn the marginal Q-value approximator and evaluate the contribution of each column in each time slot. In our experiments on two common benchmarks, CSP and VRPTW, we show that FFCG converges faster both in terms of the number of iterations and computing time compared to several greedy and ML-based baselines. Our findings also indicate that dynamically adjusting the suggested column size in each iteration is a promising approach to accelerate CG convergence.

There are several limitations of our work and we leave them for future work: (1) The reward function that we designed here still depends on hyperparameters $\alpha$ and $\beta$, which vary with the problem size. Further study can be done on automating the reward design to make it more suitable for the specific problem. 
(2) Dual stabilization, heuristic, and meta-heuristic methods can be used for further improvement. (3) When the number of variables in the RMP becomes too large, non-basic columns with the current reduced cost exceeding a given threshold may be removed. This will further reduce the computing cost of redundant columns.

\appendix

\section*{Acknowledgments}

The research is partially supported by National Key R\&D Program of China under Grant  2021ZD0110400, 
Anhui Provincial Natural Science Foundation under Grant 2208085MF172, Innovation Program for Quantum Science and Technology 2021ZD0302900 and China National Natural Science Foundation with No. 62132018, 62231015, ``Pioneer'' and ``Leading Goose''  R\&D Program of Zhejiang,  2023C01029, and 2023C01143. We also thank Sijia Zhang, Shuli Zeng, and the anonymous reviewers for their comments and helpful feedback.

\bibliography{aaai25}

\newpage
\section{CSP Formulation}

Formally, the CSP problem can be modeled by the following Integer Linear Programming (ILP):
\begin{equation}\min \sum_{i=1}^{u} y_i\end{equation}
subject to
\begin{equation}
\begin{aligned}
\sum_{j=1}^{m} w_j\xi_{ij}\leq n y_i & \ (i=1,\dots,u),\\
    \sum_{i=1}^{u}\xi_{ij}=d_j & \ (j=1,\dots,m),\\
    y_i\in\{0,1\} & \ (i=1,\dots,u),\\
     \xi_{ij}\in \mathbb{N} & \ (i=1,\dots,u;j=1,\dots,m).
\end{aligned}
 \end{equation}
where $n$ is the roll length, $m$ is the number of item types, $d_j(j=1,\dots,m)$ is the demand of item type $j$ with weight $w_j$. The objective is to find the variables $\xi_{ij}$ (integer) that satisfy these constraints and minimize the total number of larger rolls or sheets used.

In the context of CG for CSP, the SP involves generating new cutting patterns that can potentially reduce the objective function value of the master problem. The SP is typically formulated as a knapsack problem, where the goal is to find a new pattern that minimizes the reduced cost.
Given the dual variables \(\pi_j\) (associated with the demand constraints in the master problem), the SP can be formulated as follows:

\begin{equation}
\max \sum_{j=1}^{m} \pi_j \xi_j
\end{equation}
subject to

\begin{equation}
\begin{aligned}
    \sum_{j=1}^{m} w_j  \xi_j \leq n,\\
    \xi_j \in \mathbb{N}\  (j=1, \dots, m).
\end{aligned}
\end{equation}

where: \( \pi_j \) are the dual variables corresponding to the demand constraints of item type \( j \) in the master problem, \( w_j \) is the weight (or size) of item type \( j \), \( \xi_j \) represents the number of items of type \( j \) in the new pattern, \( n \) is the length of the roll (or size of the sheet).

The objective of the SP is to find a new cutting pattern (represented by \( \xi_j \)) that maximizes the total dual value \( \sum_{j=1}^{m} \pi_j \xi_j \), subject to the constraint that the total size of the items in the pattern does not exceed the roll length \( n \).

The reduced cost associated with adding a new pattern to the MP is calculated as:

\begin{equation}
\text{Reduced Cost} = 1 - \sum_{j=1}^{m} \pi_j \xi_j
\end{equation}

If the reduced cost is negative, the new pattern can potentially improve the current solution of the master problem, and it is added to the set of columns (patterns) in the MP.


\section{VRPTW Formulation}
The VRPTW problem can be formally described as the following multicommodity network flow model with time window and capacity constraints:
\begin{equation}
    \min \sum_{k\in K}\sum_{i\in N}\sum_{j\in N}c_{ij}x_{ijk}
\end{equation}
subject to
\begin{equation}
\begin{aligned} 
\sum_{k\in K}\sum_{j\in N}x_{ijk}=1,&\ \ \forall i \in C\\
\sum_{j\in N}x_{0jk}=1,&\ \ \forall k \in K\\
    \sum_{i\in N}x_{ihk}-\sum_{j\in N}x_{hjk}=0,&\ \ \forall k \in K, \forall h\in C\\
    \sum_{i\in N}x_{i,n+1,k}=1,&\ \ \forall k \in K\\
    x_{ijk}(s_{ik}+t_{ij}-s_{jk})\leq 0,&\ \ \forall k \in K,\ \forall i,j\in N\\
    \sum_{i\in C}d_i\sum_{j\in N}x_{ijk}\leq q,&\ \ \forall k \in K\\
    a_i\leq s_{ik}\leq b_i,&\ \ \forall k \in K,\forall i\in N\\
    x_{ijk}\in \{0,1\},&\ \ \forall k \in K,\forall i,j\in N
\end{aligned}
\end{equation}
where $K$ is a fleet of vehicles, $C=\{1,2,\dots,n\}$ is a set of customers, $N=C\cup\{0,n+1\}$. The depot is represented by the vertices $0$ (the
starting depot) and $n+1$ (the returning depot). And $x_{ijk}=1$, if  vehicle $k$ drives directly from vertex $i$ to vertex $j$. Otherwise, $x_{ijk}=0$. The decision variable $s_{ik}$ is defined for each vertex $i$ and each vehicle $k$ and denotes the time vehicle
$k$ starts to service customer $i$. In case of vehicle $k$ does not service customer $i$, $s_{ik}$ has no meaning and
consequently its value is considered irrelevant. We assume $a_0 = 0$ and therefore $s_{0k} = 0$, for all $k$.

In CG for the VRPTW, the SP decomposes into $|K|$ identical problems, each one being an Elementary Shortest Path Problem with Resource Constraints (ESPPRC).

\section{Computing Environment}
We implement our FFCG in Python based on Tensorflow 2.12.0. To solve both RMP and SP optimization problems in CG, we use Gurobi 11.0.2 \cite{gurobi}. For the training using 440 instances scheduled for CSP and 240 instances scheduled for VRPTW, the training takes around 15-21 hours CPU time using AMD EPYC 7763 64-Core Processor @ 2.45GHz.

\section{Node Features}
In this section, we describe node features used for CSP and VRPTW:
\subsection{Column node features}
Each column node corresponds to one decision variable. Features (1) and (2) relate to solving the RMP problem as they are all information about decision variables
in RMP. Feature (3)  is determined by
the problem formulation of each problem instance, while features (5) - (8) correspond to the dynamical
information of each column entering and leaving the basis. Feature (9) is a problem-specific feature, which can be designed to get the characteristics of different problems.
\begin{enumerate}
    \item \textbf{Reduced cost}: Reduced cost is a quantity associated with each variable indicating how much the objective
function coefficient on the corresponding variable must be improved before the solution value of the
decision variable will be positive in the optimal solution. The reduced cost value is only non-zero when the optimal value of a variable is zero.

\item \textbf{Solution value}: The solution value of each decision variable corresponding to each column node after
solving the RMP in the current iteration. For each existing column node, this feature is a continuous number greater
than or equal to 0. The candidate column nodes have this feature set to be 0.
\item \textbf{Connectivity of column node}: Total number of constraint nodes each column node connects to. In CSP, as each
constraint is a particular demand, this node feature indicates the ability of a column node (a pattern) to
satisfy demands. It also indicates the connectivity of each column node in the bipartite graph representing
the state.

\item \textbf{Number of iterations that each column node stays in the basis}: If the column node stays in the basis
for a long time, it is most likely that the pattern corresponds to this column node is really good.
\item \textbf{Number of iterations that each column node stays out of the basis}: if the column node stays out of
the basis for a long time, it is most likely never entering the basis and being used in optimal solution in
future iterations.
\item \textbf{If the column left basis on the last iteration or not}: This is a binary feature recording the dynamics of
each column node.
\item \textbf{If the column enter basis on the last iteration or not}: Similar binary feature as (6).
\item \textbf{Node status}: A feature indicating a column node is a candidate (to be selected), suggested (selected), or existing node. If the column node is a candidate node (column) that is generated at the current iteration by SP, then this feature is 1 (to be selected) or 0 (selected), otherwise -1 (existing).
\item \textbf{Problem-specific feature}: In CSP, it is Waste, a feature recording the remaining length of a roll if we were to cut the current pattern from the
roll. Again, each column node corresponds to one decision variable in RMP, which also represents one
particular cut pattern. In VRPTW, it is Route cost, a feature recording the cost of each route. It also corresponds to one decision variable in RMP, which represents one possible route.
\end{enumerate}

\subsection{Constraint node features}
Each constraint node corresponds to one constraint in RMP, so the number of constraint nodes is fixed for
each problem instance.
\begin{enumerate}
    \item \textbf{Dual value}: Dual value or shadow price is the coefficient of each dual variable in SP objective
function, and as each constraint node corresponds to one dual variable, we record dual value as one
feature for constraint node.
\item \textbf{Connectivity of constraint node}: Total number of column nodes each constraint node connects to,
which also indicates the connectivity of each constraint node in the bipartite graph representing the state.
\end{enumerate}

\section{Hyperparameter Tuning}
We conduct hyperparameter tuning using a validation set for CSP, which includes 30 instances  with the roll length varying from 50 to 200. The values we consider for each hyperparameter are the following: $\alpha \in \{100,500,1000,2000,3000\}$, $\beta\in\{0.1,0.3,0.5,0.7,0.9\}$. Increasing the values of parameters $\alpha$  will enable the RL agent to choose more columns. On the other hand, a higher value of $\beta$ will prevent the RL agent from selecting too many unnecessary columns. The search space for hyperparameters is defined as the Cartesian product between all these sets of different hyperparameters possible values, which gives us 25 configurations and we randomly select 14 configurations out of them.
In Table \ref{tab:valid}, we provide detailed configurations of each model index for our analysis of hyperparameters as well as their detailed validation results. The best
configuration is model 9: $\alpha= 2000$, and $\beta= 0.3$. For all the results reported
in this paper, we use this configuration.

\begin{table}[tb]
\centering

\begin{tabular}{crr}
\toprule
Model index &  $(\alpha,\beta)$  & Runtime (s)           \\ \midrule
1   & $(100,0.5)$ &  $41.8$\\
2  & $(100,0.9)$ & $44.2$\\
3  & $(500,0.1)$ &  $36.4$\\
4  & $(500,0.3)$ &  $32.5$\\
5  & $(500,0.9)$ &  $45.6$\\
6  & $(1000,0.3)$ &  $38.7$\\
7  & $(1000,0.5)$ & $45.7$ \\
8  & $(1000,0.7)$ & $41.8$\\
9  & $(2000,0.3)$ & $25.4$ \\
10  & $(2000,0.5)$ &  $27.3$\\
11 & $(2000,0.7)$ &  $29.8$\\
12 & $(3000,0.1)$ & $27.6$ \\
13  & $(3000,0.5)$ & $29.1$ \\
14  & $(3000,0.7)$ &  $36.1$\\
\bottomrule
\end{tabular}
\caption{Model configurations and validation performances.}\label{tab:valid}
\end{table}

\section{Dataset}
\begin{table*}[tb]
    \centering
    \begin{tabular}{c|c|c|c|c|c|c} \toprule 
 Dataset&  Total&$n=50$&  $n=100$&$n=200$& $n=750$& $n=1000$\\ \midrule
         Training&  400  &160&  160&80& 0& 0\\  
         Validation& 
       30&10&  10&10& 0& 0\\ 
 Testing&  265&182&  0&46& 15& 22\\\bottomrule\end{tabular}
    \caption{Dataset division for CSP}
    \label{tab:dd}
\end{table*}

\subsection{Cutting Stock Problem}
We use BPPLIB \cite{bpplib}, a widely used benchmark for binary packing and cutting stock problems, which provides over 6000 instances. BPPLIB contains instances of different sizes with the roll length $n$ varying from 50 to 1000 and the number of item types $m$ fluctuating from 20 to 500. The cutting patterns (variables) size is $2^m$.
Table \ref{tab:dd}  shows the information of the instances contained in the training set, validation set, and
testing set for CSP. Column ``Total'' lists the total number of instances in each dataset, while other columns list the number of instances with specific roll length $n$ in that dataset.

\subsection{Vehicle Routing Problem with Time Windows}
We use Solomon benchmark \cite{Solomon1987AlgorithmsFT} for VRPTW. This dataset contains six different problem types (C1, C2, R1, R2, RC1, RC2), each of which has 8-12 instances with 50 customers. ``C'' refers to customers who are geographically clustered, ``R'' to randomly placed customers, ``RC'' to a mixture. The ``1'' and ``2'' labels refer to narrow time windows/small vehicle capacity and large time windows/large vehicle capacity, respectively. The difficulty levels of these sets are in order of C, R, RC. There are 56 instances in Solomon’s dataset. We generate smaller instances by considering only the first $n$ customers, where $n$ is randomly sampled from 5–16, customers from each original Solomon instance.

We use instances from types C1, R1, and RC1 for training. For the training set, we generate 80 smaller instances per type from the original Solomon's instances for a total of 240 training instances. For testing, we consider 60 other larger instances from types C1, R1, and RC1.

\begin{figure}[tb]
    \centering
\includegraphics[width=0.35\textwidth]{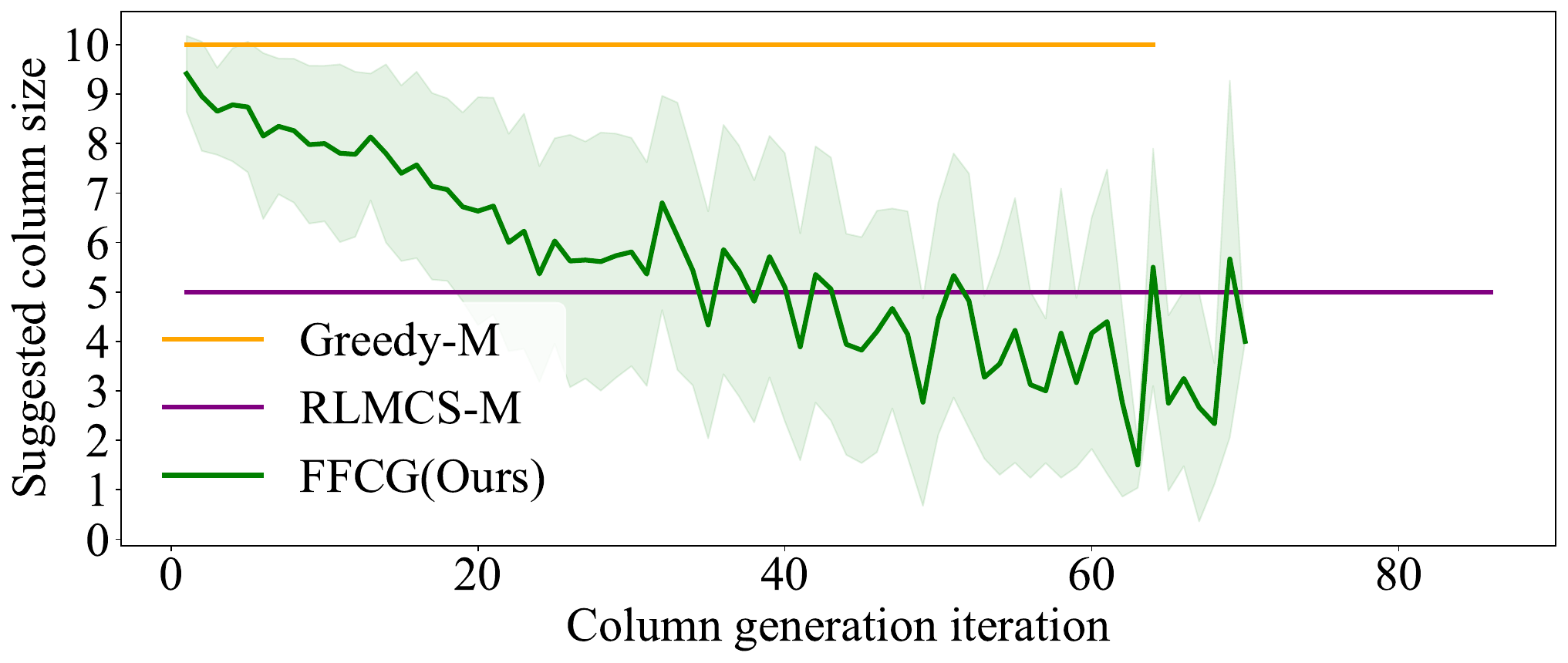}
    \caption{CSP, $n=200$ test instances: The suggested column size per CG iteration for FFCG, Greedy-M, and RLMCS-M.}
    \label{fig:Analysis200}
\end{figure}

\section{Detailed Statistics of Testing Results}

\begin{figure}[tb]
    \centering
    \includegraphics[width=0.35\textwidth]{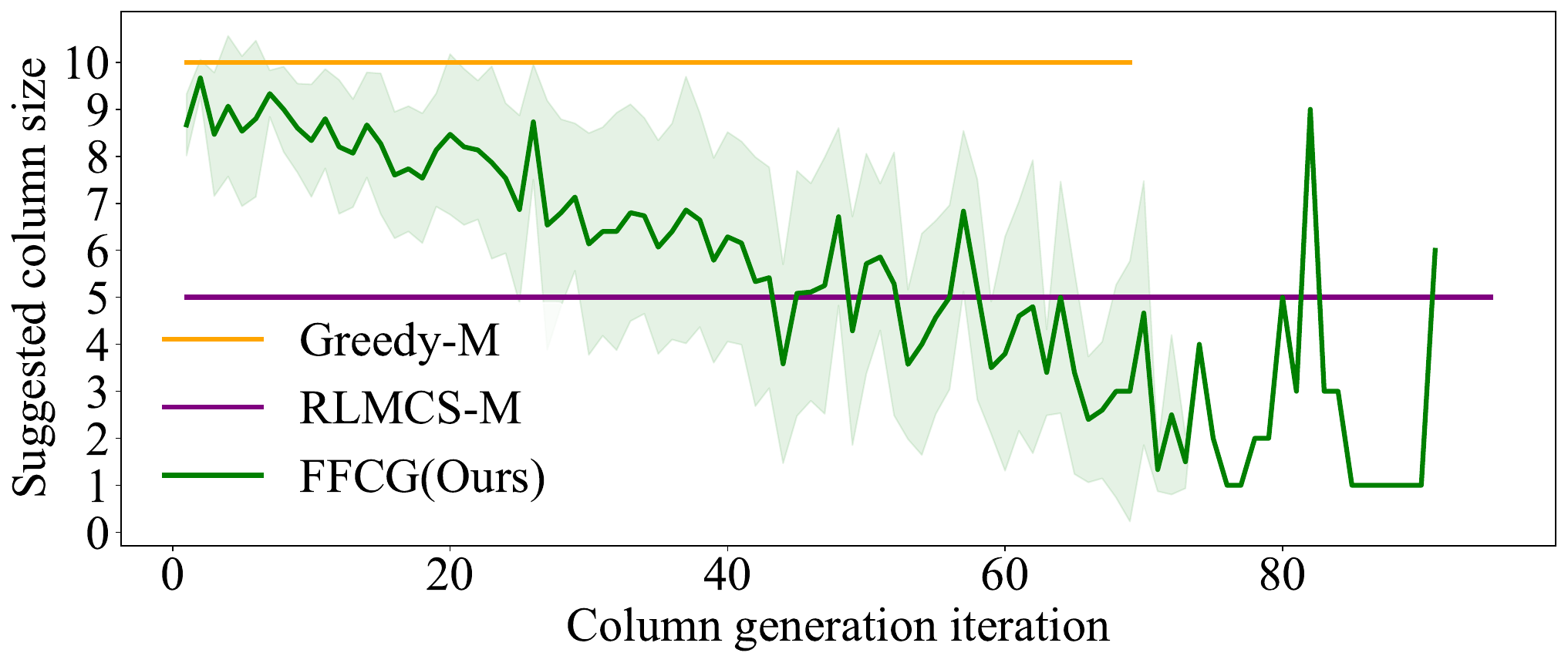}
    \caption{CSP, $n=750$ test instances: The suggested column size per CG iteration for FFCG, Greedy-M, and RLMCS-M.}
    \label{fig:Analysis750}
\end{figure}
\begin{figure}[htb]
    \centering
\includegraphics[width=0.35\textwidth]{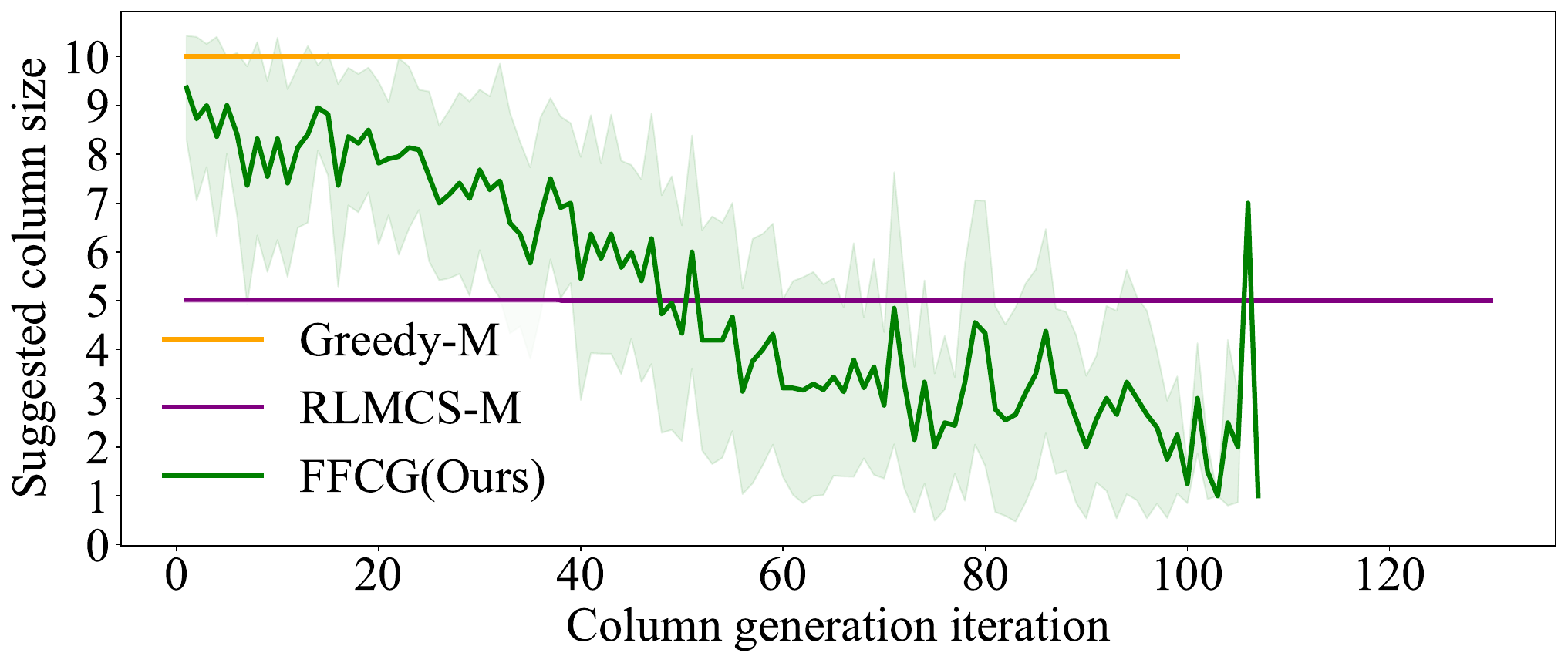}
    \caption{CSP, $n=1000$ test instances: The suggested column size per CG iteration for FFCG, Greedy-M, and RLMCS-M.}
    \label{fig:Analysis1000}
\end{figure}

In Table \ref{tab:Detailed50}, Table \ref{tab:Detailed200}, Table \ref{tab:Detailed750} and Table \ref{tab:Detailed1000}, we present statistics obtained from our experimental results for CSP. We report the average and standard deviation of the number of iterations, solution time (measured in seconds), the number of selected columns, and
the objective function values.  Our findings clearly indicate the superiority of FFCG over other approaches in effectively solving complex CG problems in practice. 
We further analyze the suggested column size for other test cases. As shown in Figure \ref{fig:Analysis200}, Figure \ref{fig:Analysis750}, and Figure \ref{fig:Analysis1000}, FFCG follows the same strategy and reduces the suggested column size as CG processes.

\section{VRPTW Convergence Plots}

\begin{figure}[htb]
    \centering
    \includegraphics[width=0.35\textwidth]{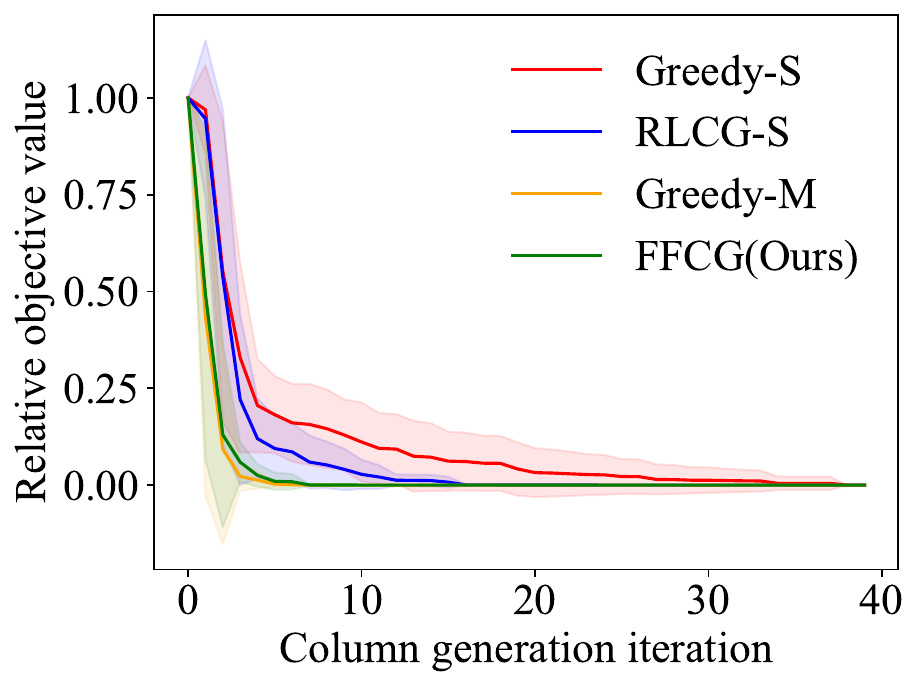}
    \caption{VRPTW: CG convergence plots for FFCG, Greedy-S, RLCG-S, and Greedy-M.}
    \label{fig:vrpe}
\end{figure}

In Figure \ref{fig:vrpe}, we illustrate the CG convergence speed of different strategies on VRPTW testing cases.  The objective values are normalized to the range $[0,1]$ before averaging across instances.

\begin{table*}[tb]
\begin{center}
\begin{tabular}{c|rrrrrrrr} \toprule
 \multirow{2}{*}{Strategy}&  \multicolumn{2}{c|}{\#Itr} &  \multicolumn{2}{c|}{Time} &    \multicolumn{2}{c|}{\#Col}&\multicolumn{2}{c}{ObjVal}  \\
 & \multicolumn{1}{|c}{$\mu$}& \multicolumn{1}{c|}{$\sigma$}& \multicolumn{1}{c}{$\mu$}& \multicolumn{1}{c|}{$\sigma$}& \multicolumn{1}{c}{$\mu$}& \multicolumn{1}{c|}{$\sigma$}& \multicolumn{1}{c}{$\mu$}& \multicolumn{1}{c}{$\sigma$}\\ \midrule
Greedy-S &  53.10&\multicolumn{1}{c|}{16.44}&  4.19&\multicolumn{1}{c|}{2.23}&   53.10&\multicolumn{1}{c|}{16.44}&25.84&  2.83\\
MLCG-S &    43.48&\multicolumn{1}{c|}{13.57}&  5.10&\multicolumn{1}{c|}{2.66}&   43.48&\multicolumn{1}{c|}{13.57}&25.84&  2.83\\
RLCG-S &  43.21&\multicolumn{1}{c|}{14.31}&  3.76&\multicolumn{1}{c|}{1.95}&   43.21&\multicolumn{1}{c|}{14.31}&25.84&  2.83\\\midrule
Greedy-M &   11.80&\multicolumn{1}{c|}{3.09}&  1.34&\multicolumn{1}{c|}{0.68}&   117.97&\multicolumn{1}{c|}{30.90}&25.84&  2.83\\
RLMCS-M &  13.87&\multicolumn{1}{c|}{4.00}&  1.69&\multicolumn{1}{c|}{0.46}&   69.37&\multicolumn{1}{c|}{20.00}&25.84&  2.83\\
FFCG &   11.81&\multicolumn{1}{c|}{3.28}&  1.30&\multicolumn{1}{c|}{0.66}&   77.99&\multicolumn{1}{c|}{21.29} &25.84& 2.83\\\bottomrule
\end{tabular}
\caption{CSP, $n=50$ instances: Iterations, solution time (in seconds), selected columns, and objective value reports with $\mu$ mean, and $\sigma$ standard
deviation.}
    \label{tab:Detailed50}
\end{center}
\end{table*}

\begin{table*}[tb]
\begin{center}
\begin{tabular}{c|rrrrrrrr} \toprule
 \multirow{2}{*}{Strategy}&  \multicolumn{2}{c|}{\#Itr} &  \multicolumn{2}{c|}{Time} &    \multicolumn{2}{c|}{\#Col}&\multicolumn{2}{c}{ObjVal}  \\
 & \multicolumn{1}{|c}{$\mu$}& \multicolumn{1}{c|}{$\sigma$}& \multicolumn{1}{c}{$\mu$}& \multicolumn{1}{c|}{$\sigma$}& \multicolumn{1}{c}{$\mu$}& \multicolumn{1}{c|}{$\sigma$}& \multicolumn{1}{c}{$\mu$}& \multicolumn{1}{c}{$\sigma$}\\ \midrule
Greedy-S &  147.30&\multicolumn{1}{r|}{56.40}&  9.47&\multicolumn{1}{r|}{4.28}&   147.30&\multicolumn{1}{r|}{56.40}&102.39&  6.86\\
MLCG-S &    145.74&\multicolumn{1}{r|}{54.22}&  21.78&\multicolumn{1}{r|}{18.63}&   145.74&\multicolumn{1}{r|}{54.22}&102.39&  6.86\\
RLCG-S &  152.80&\multicolumn{1}{r|}{58.14}&  8.77&\multicolumn{1}{r|}{5.33}&   152.80&\multicolumn{1}{r|}{58.14}&102.39&  6.86\\\midrule
Greedy-M &   35.15&\multicolumn{1}{r|}{13.68}&  2.69&\multicolumn{1}{r|}{1.82}&   351.52&\multicolumn{1}{r|}{136.76}&102.39&  6.86\\
RLMCS-M &  45.67&\multicolumn{1}{r|}{17.16}&  5.59&\multicolumn{1}{r|}{2.38}&   228.37&\multicolumn{1}{r|}{85.8}&102.39&  6.86\\
FFCG &   38.57&\multicolumn{1}{r|}{16.38}&  2.51&\multicolumn{1}{r|}{1.65}&   250.63&\multicolumn{1}{r|}{92.91} &102.39& 6.86\\\bottomrule
\end{tabular}
\caption{CSP, $n=200$ instances: Iterations, solution time (in seconds), selected columns, and objective value reports with $\mu$ mean, and $\sigma$ standard
deviation.}
    \label{tab:Detailed200}
\end{center}
\end{table*}

\begin{table*}[tb]
\begin{center}
\begin{tabular}{c|rrrrrrrr} \toprule
 \multirow{2}{*}{Strategy}&  \multicolumn{2}{c|}{\#Itr} &  \multicolumn{2}{c|}{Time} &    \multicolumn{2}{c|}{\#Col}&\multicolumn{2}{c}{ObjVal}  \\
 & \multicolumn{1}{|c}{$\mu$}& \multicolumn{1}{c|}{$\sigma$}& \multicolumn{1}{c}{$\mu$}& \multicolumn{1}{c|}{$\sigma$}& \multicolumn{1}{c}{$\mu$}& \multicolumn{1}{c|}{$\sigma$}& \multicolumn{1}{c}{$\mu$}& \multicolumn{1}{c}{$\sigma$}\\ \midrule
Greedy-S &  222.20&\multicolumn{1}{r|}{56.61}&  16.96&\multicolumn{1}{r|}{7.88}&   222.20&\multicolumn{1}{r|}{56.61}&383.53&  12.11\\
MLCG-S &    232.93&\multicolumn{1}{r|}{58.31}&  31.93&\multicolumn{1}{r|}{21.40}&   232.93&\multicolumn{1}{r|}{58.31}&383.53&  12.11\\
RLCG-S &  237.67&\multicolumn{1}{r|}{62.29}&  16.10&\multicolumn{1}{r|}{7.18}&   237.67&\multicolumn{1}{r|}{62.29}&383.53&  12.11\\\midrule
Greedy-M &   51.13&\multicolumn{1}{r|}{12.52}&  5.38&\multicolumn{1}{r|}{2.57}&   511.33&\multicolumn{1}{r|}{125.16}&383.53&  12.11\\
RLMCS-M &  71.33&\multicolumn{1}{r|}{16.73}&  12.93&\multicolumn{1}{r|}{3.49}&   356.67&\multicolumn{1}{r|}{83.66}&383.53&  12.11\\
FFCG &   56.27&\multicolumn{1}{r|}{15.12}&  4.50&\multicolumn{1}{r|}{2.28}&   377.40&\multicolumn{1}{r|}{84.36} &383.53& 12.11\\\bottomrule
\end{tabular}
\caption{CSP, $n=750$ instances: Iterations, solution time (in seconds), selected columns, and objective value reports with $\mu$ mean, and $\sigma$ standard
deviation.}
    \label{tab:Detailed750}
\end{center}
\end{table*}

\begin{table*}[tb]
\begin{center}
\begin{tabular}{c|rrrrrrrr} \toprule
 \multirow{2}{*}{Strategy}&  \multicolumn{2}{c|}{\#Itr} &  \multicolumn{2}{c|}{Time} &    \multicolumn{2}{c|}{\#Col}&\multicolumn{2}{c}{ObjVal}  \\
 & \multicolumn{1}{|c}{$\mu$}& \multicolumn{1}{c|}{$\sigma$}& \multicolumn{1}{c}{$\mu$}& \multicolumn{1}{c|}{$\sigma$}& \multicolumn{1}{c}{$\mu$}& \multicolumn{1}{c|}{$\sigma$}& \multicolumn{1}{c}{$\mu$}& \multicolumn{1}{c}{$\sigma$}\\ \midrule
Greedy-S &  386.14&\multicolumn{1}{r|}{103.57}&  67.34&\multicolumn{1}{r|}{48.62}&   386.14&\multicolumn{1}{r|}{103.57}&113.50&  21.88\\
MLCG-S &    295.09&\multicolumn{1}{r|}{71.92}&  70.89&\multicolumn{1}{r|}{46.61}&   295.09&\multicolumn{1}{r|}{71.92}&113.50&  21.88\\
RLCG-S &  300.86&\multicolumn{1}{r|}{75.52}&  46.88&\multicolumn{1}{r|}{32.11}&   300.86&\multicolumn{1}{r|}{75.52}&113.50&  21.88\\\midrule
Greedy-M &   66.45&\multicolumn{1}{r|}{15.03}&  11.51&\multicolumn{1}{r|}{8.07}&   664.55&\multicolumn{1}{r|}{150.32}&113.50&  21.88\\
RLMCS-M &  86.50&\multicolumn{1}{r|}{18.23}&  17.80&\multicolumn{1}{r|}{5.45}&   432.50&\multicolumn{1}{r|}{91.14}&113.50&  21.88\\
FFCG &   78.86&\multicolumn{1}{r|}{17.13}&  9.65&\multicolumn{1}{r|}{4.93}&   462.73&\multicolumn{1}{r|}{88.69} &113.50& 21.88\\\bottomrule
\end{tabular}
\caption{CSP, $n=1000$ instances: Iterations, solution time (in seconds), selected columns, and objective value reports with $\mu$ mean, and $\sigma$ standard
deviation.}
    \label{tab:Detailed1000}
\end{center}
\end{table*}

\end{document}